\documentclass[letterpaper]{article} 
\usepackage{aaai2026}  
\usepackage{times}  
\usepackage{helvet}  
\usepackage{courier}  
\usepackage[hyphens]{url}  
\usepackage{graphicx} 
\usepackage{natbib}  
\usepackage{caption} 
\usepackage{algorithm}
\usepackage{algorithmic}

\usepackage{amssymb}
\usepackage{booktabs}
\usepackage{booktabs}
\usepackage{multirow}
\usepackage{multirow}
\usepackage{booktabs}
\usepackage{pifont}       
\usepackage{bbding}       
\usepackage{fontawesome}  

\usepackage{array}      
\usepackage{tabularx}   
\usepackage{booktabs}   
\usepackage{multirow}   
\usepackage{caption}    
\usepackage{makecell}   
 
\usepackage{amsmath}

\usepackage{newfloat}
\usepackage{listings}
\DeclareCaptionStyle{ruled}{labelfont=normalfont,labelsep=colon,strut=off} 
\floatstyle{ruled}
\newfloat{listing}{tb}{lst}{}
\floatname{listing}{Listing}
\title{AFR-CLIP: Enhancing Zero-Shot Industrial Anomaly Detection with Stateless-to-Stateful Anomaly Feature Rectification}
\author{
    Jingyi Yuan\textsuperscript{\rm 1}, Chenqiang Gao\textsuperscript{\rm 1}, Pengyu Jie\textsuperscript{\rm 1}, Xuan Xia\textsuperscript{\rm 2}, Shangri Huang\textsuperscript{\rm 1}, Wanquan Liu\textsuperscript{\rm 1}
}
\affiliations{
    \textsuperscript{\rm 1}School of Intelligent Systems Engineering, Sun Yat-Sen University\\
    \textsuperscript{\rm 2}Shenzhen Institute of Artificial Intelligence and Robotics for Society, Shenzhen\\
    \{yuanjy36, jiepyu3, huangshr25\}@mail2.sysu.edu.cn, \{gaochq6,liuwq63\}@mail.sysu.edu.cn, xiaxuan@cuhk.edu.cn.

}

\usepackage{bibentry}

\begin{document}

\maketitle

\begin{abstract}
Recently, zero-shot anomaly detection (ZSAD) has emerged as a pivotal paradigm for industrial inspection and medical diagnostics, detecting defects in novel objects without requiring any target-dataset samples during training. 
Existing CLIP-based ZSAD methods generate anomaly maps by measuring the cosine similarity between visual and textual features. 
However, CLIP's alignment with object categories instead of their anomalous states limits its effectiveness for anomaly detection. To address this limitation, we propose AFR-CLIP, a CLIP-based anomaly feature rectification framework.
AFR-CLIP first performs image-guided textual rectification, embedding the implicit defect information from the image into a stateless prompt that describes the object category without indicating any anomalous state. 
The enriched textual embeddings are then compared with two pre-defined stateful (normal or abnormal) embeddings, and their text-on-text similarity yields the anomaly map that highlights defective regions.
To further enhance perception to multi-scale features and complex anomalies, we introduce self prompting (SP) and multi-patch feature aggregation (MPFA) modules.
Extensive experiments are conducted on eleven anomaly detection benchmarks across industrial and medical domains, demonstrating AFR-CLIP's superiority in ZSAD.
\end{abstract}


\section{Introduction}
\label{sec:intro}
\begin{figure}[!h]
	\centering
	\includegraphics[width=0.48 \textwidth]{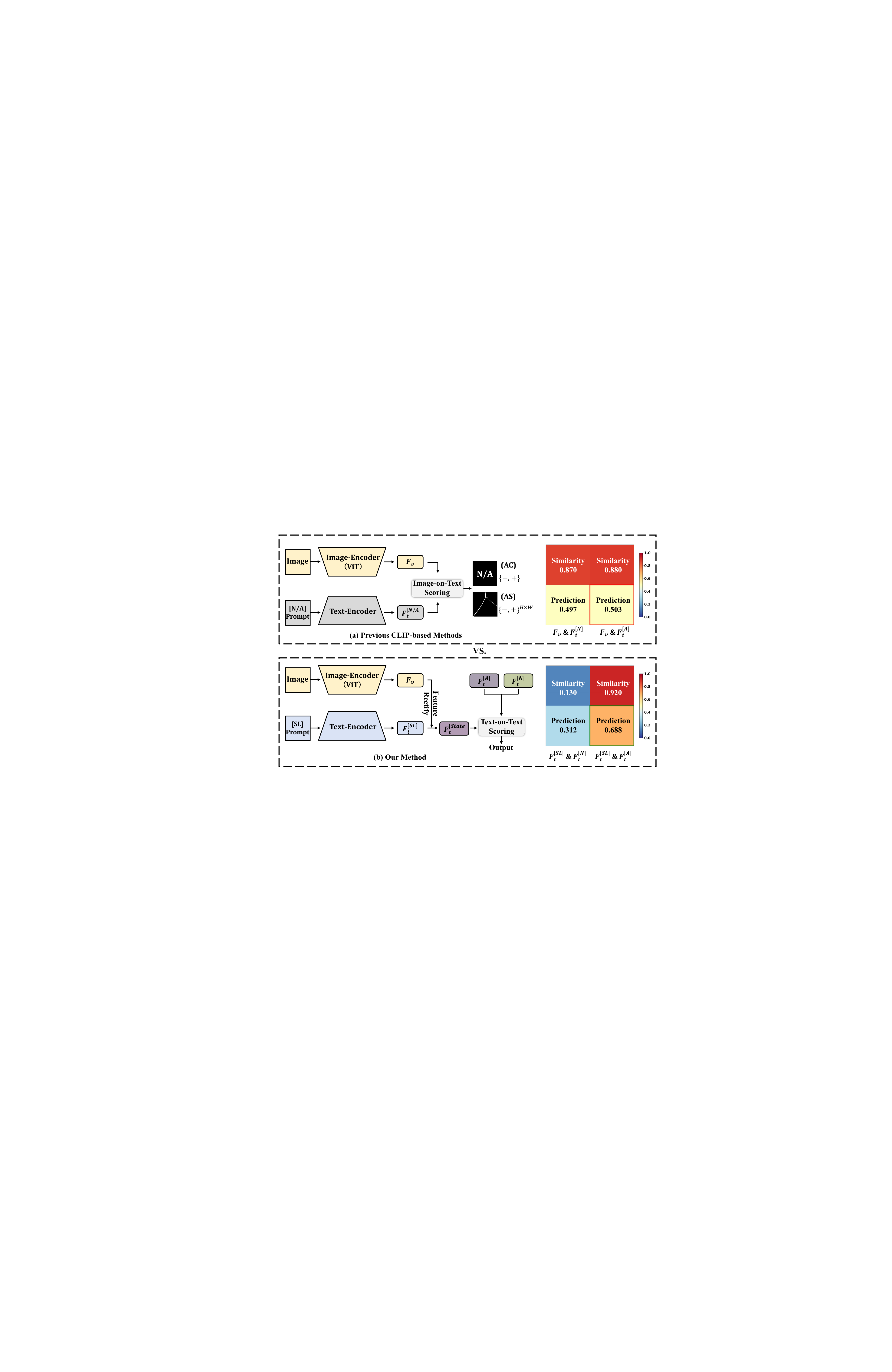}%
	\caption{\textbf{Comparison between previous CLIP-based methods and our method.} (a) Previous CLIP-based methods obtain the mask directly by computing the similarity between visual and textual features, failing to detect anomalous states effectively, as reflected by the almost indistinguishable similarity scores. This is because CLIP's alignment during pretraining stage is dominated by the category semantics. (b) Our method with anomaly rectification module calibrates defect cues into a stateless (SL) prompt, explicitly suppressing category semantics and elevating anomaly-state signals to enlarge the similarity margin between the two states.}
	\label{fig:previous}
\end{figure}
Anomaly detection (AD) holds substantial potential across a wide range of domains, including industrial product quality control, medical diagnostics, etc. Industrial anomaly detection (IAD) focuses on identifying atypical patterns in industrial images. Owing to the scarcity of abnormal samples and the high cost of data annotation, numerous unsupervised anomaly detection methods have emerged~\cite{un01,un02,un03,un04,un05,un06,un07,patchcore,spade,padim}. These methods detect and locate defects by measuring deviations from the distribution of normal data.
Nevertheless, because they model the distribution only for the categories on training datasets, their performance degrades sharply when confronted with the wide variety of objects and defect types in real-world industrial domain.

Since the unsupervised methods need to re-model the normal distribution when a new product category appears, recent works have increasingly focused on zero-shot anomaly detection (ZSAD). ZSAD aims to detect anomalies in novel objects without any target-dataset samples during training. 
Many existing ZSAD works exploit pre-trained vision–language models (VLMs) for their strong generalization ability~\cite{anomalyclip,winclip,adaclip,aprilgan,cocoop,coop,clipad}. Owing to the remarkable image-text alignment and representation capability of  CLIP~\cite{clip}, a series of CLIP-based ZSAD methods~\cite{adaclip,anomalyclip,clipad,denseclip,cocoop} compute cosine similarity between visual features and text prompts as the anomaly map, shown in Fig. \ref{fig:previous}(a). Despite these efforts, one core issue remains:
The similarities between the image and the stateful text prompts (e.g., “a photo of a normal/abnormal screw”) are almost indistinguishable, often centered around 0.5, revealing the model’s insensitivity to state semantics.
Due to the lack of explicit anomaly annotations during CLIP’s pretraining, its ability to encode fine-grained abnormal regions is fundamentally limited. This is further compounded by CLIP’s alignment mechanism, which tends to focus on object categories (e.g., screw) rather than their anomalous states (e.g., normal vs. abnormal).



To address the aforementioned problems, we propose AFR-CLIP, a novel framework that fundamentally shifts the focus from object categories to anomalous states by introducing a cross-modal feature rectification strategy.
Unlike traditional methods shown in Fig. \ref{fig:previous}(a) that rely solely on image-text similarity, our approach integrates image-specific defect information directly into the text embeddings, enabling more precise anomaly detection and localization in zero-shot settings (Fig. \ref{fig:previous}(b)).
Since CLIP’s pre-training already aligns image features with textual object category, the rectification process can focus exclusively on injecting the anomalous state information.
Instead of comparing visual and textual features, we then compute text-on-text similarity between the rectified embedding and two universal stateful prototypes (normal and abnormal), ensuring that the similarity scores respond to anomalous state rather than mere object category.
Thus, the injected anomalous state cues dominate the similarity computation, allowing AFR-CLIP to segment defective regions without being confounded by category semantics.
Additionally, we equip AFR-CLIP with two lightweight modules, self prompting (SP) and multi-patch feature aggregation (MPFA), which aggregate global and local features in multi-scale to enhance fine-grained defects localization.


The main contributions are summarized as follows:

$\bullet$ We propose AFR-CLIP, a novel zero-shot anomaly detection framework that embeds image-specific defect cues into stateless prompts and performs text-on-text scoring against stateful prototypes to enhance anomaly localization.

$\bullet$ We design SP and MPFA modules to fuse local-global context and improve fine-grained and multi-scale detection.

$\bullet$ We validate AFR-CLIP on eleven datasets across industrial and medical domains, achieving state-of-the-art performance in both detection and localization tasks.

\section{Related Work}
\label{sec:related work}
Abnormal samples in industrial and medical domains are scarce and costly, so it is hard to collect plenty of defective images for training. Many methods explore in the zero-shot setting.



\subsection{Prompt Learning}
As an emerging technology in the field of natural language processing (NLP), prompt learning has received extensive attention in recent years. The core idea is to design appropriate prompts to guide the pre-trained model to adapt to a specific task, without the need for a large amount of task-specific annotated data. With the deepening of research, prompt learning has gradually expanded from the single text tasks to the multi-modal settings. For instance, CoOp \cite{coop} employs learnable prompt tokens into text branch, first introducing prompt learning in CLIP. Subsequently, DenseCLIP \cite{denseclip} and CoCoOp \cite{cocoop} further exploit visual context prompting to adapt VLMs to the target domain. Building on these advances, we propose self prompting mechanism to refine global semantics with local features, better localizing small and complex anomalies.

\subsection{Zero-Shot Anomaly Detection}
\label{sec:zsad}
Recently, visual-language pre-trained models such as MiniGPT-4 \cite{minigpt4}, LLAVA \cite{LLAVA}, Otter \cite{otter} and CLIP \cite{clip} present promising strong abilities in the zero-shot settings. Especially, CLIP is one popular backbone for ZSAD, which aims to embed visual and textual features into a shared space. Zero-Shot Anomaly Detection methods \cite{adaclip,anomalyclip,aprilgan,clipad,cocoop,coop,denseclip,winclip} often utilize a few auxiliary seen objects and anomalies during training to detect novel categories, relying on remarkable generalization capability of VLMs. For the strong image-text alignment ability, many existing ZSAD methods select CLIP~\cite{clip} as the baseline. In particular, WinCLIP \cite{winclip} designs normal and abnormal text prompts manually, and then computes the cosine similarity between visual and textual features, subsequently followed by an interpolation module to obtain anomaly map in the original resolution. In contrast to the training-free method, several approaches \cite{anomalyclip,clipad,aprilgan} train lightweight adapters with labeled data to close the domain gap between natural and task-specific domains. However, considering the expensive cost and low efficiency of handcrafted prompts, AnomalyCLIP \cite{anomalyclip} replaces handcrafted texts with two learnable prompt tokens.
Unlike existing ZSAD approaches that score anomalies via image–text cosine similarity, our AFR-CLIP first injects visual anomalous information into the stateless prompt and then matches this image-calibrated textual embedding against two universal stateful prototypes (normal and abnormal).  
This text-on-text comparison amplifies the model’s sensitivity to anomalous states while still preserving CLIP’s category alignment, enabling finer and more robust detection accuracy.

\begin{figure*}[!h]
	\centering
	\includegraphics[width=1\textwidth]{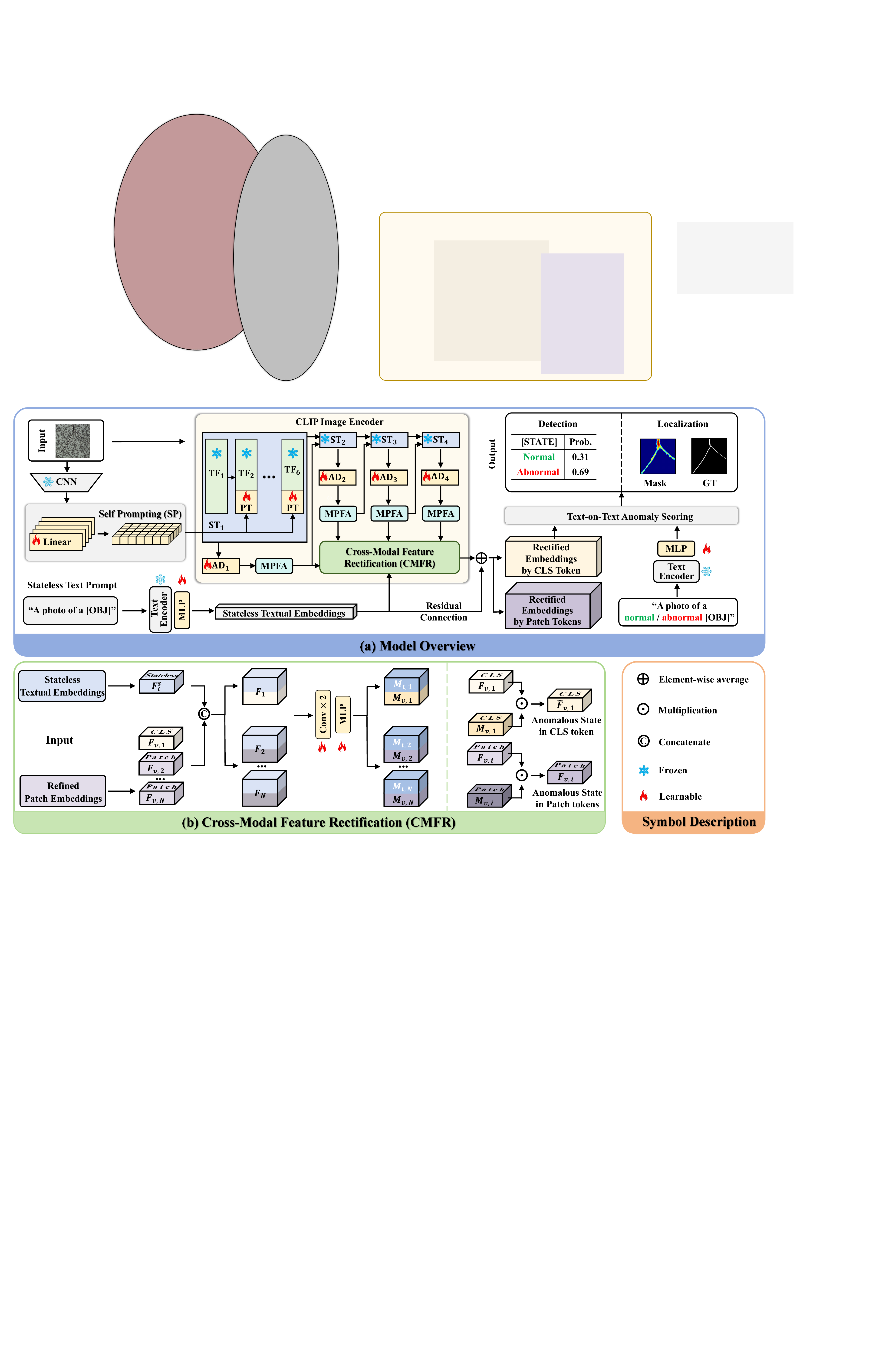} %
	\caption{Framework of AFR-CLIP.  
(a) Model Overview and (b) the detail of Cross-Modal Feature Rectification.
\textbf{Cross-Modal Feature Rectification (CMFR)} first injects patch-level visual cues into a stateless prompt, generating a state-aware text embedding that preserves category semantics while highlighting defect signals.  
This rectified embedding is then evaluated by \textbf{Text-on-Text Anomaly Scoring}, which measures the similarity with two universal prototypes (normal and abnormal) and produces image- and patch-level anomaly scores.  
Complementary modules further enhance the representation: \textbf{Self-Prompting (SP) }fuses CNN-derived local features with ViT tokens to enrich texture detail, and \textbf{Multi-Patch Feature Aggregation (MPFA)} pools information over the \(m\times m\) neighborhood to improve boundary coherence and multi-scale context.}
	\label{fig:Agg-CLIP}
\end{figure*}
\section{Method}
\label{sec:method}
\subsection{Problem Definition}
Given one test sample $I\in\mathbb{R}^{H \times W \times 3}$, zero-shot anomaly detection methods aim to produce image-level anomaly score $S\in [0,1]$ and pixel-level anomaly map $M\in [0, 1]^{H\times W}$. Following existing ZSAD methods, we leverage an auxiliary dataset $D_a = \left\{ (I_1,G_1),\ldots,(I_n,G_n)\mid G_i\in [0,1]^{H \times W} \right\}$ for training, which consists of both normal and abnormal samples $\{I\}^n_{i=1}$ with corresponding annotations $\{G\}_{i=1}^n$. Then we test on unseen dataset $D_t = \left\{ I_{t_1},\ldots,I_{t_n} \right\}$. Notably, the auxiliary dataset and test dataset originate from different domains, ensuring that $D_a \cap D_t = \emptyset$. 

\subsection{Overview}
We adopt CLIP as the backbone of our AFR-CLIP. As illustrated in Fig. \ref{fig:Agg-CLIP}(a), in the image branch, an input image is processed through CLIP's image encoder to obtain patch embeddings. The SP module refines these embeddings by fusing global and local features from CNN and ViT, respectively. These refined features are then projected linearly and processed by the MPFA module, which aggregates multi-scale features of neighboring patches.
In the text branch, we embed a text prompt (e.g., “a photo of a \{screw\}”) without any anomalous state. The textual features are followed by a linear projection layer for domain adaptation. Then, the textual embeddings are rectified by the refined visual features in the CMFR module. Finally, we perform the text-on-text anomaly scoring to obtain the prediction.

\subsection{Cross-Modal Feature Rectification (CMFR)}
In original CLIP, visual and textual features are aligned in a shared embedding space, but this alignment primarily captures object semantics (i.e., the category of the object) rather than the anomalous state (e.g., whether it is normal or defective). To address this limitation, we introduce a stateless-to-stateful anomaly features rectification that adjusts the text features with visual features.
Instead of computing image-text cosine similarity used in previous methods \cite{winclip,clipad}, we introduce text-on-text similarity. In particular, we first rectify the stateless textual embeddings with visual features to capture the hidden anomalous state information in the image. Then we measure cosine similarity between the rectified textual features and two textual features embedded from normal and abnormal prompts.

As illustrated in Fig. \ref{fig:Agg-CLIP}(b), the process begins by extracting the image feature $F_v\in \mathbb{R}^{N\times D}$ from the image encoder of CLIP.
To reduce the domain gaps, linear adapters are followed by the visual feature.
Simultaneously, three different types of text prompts $T^{\{n,a,s\}}$ are embedded into $F_t\in \mathbb{R}^{3\times D}$ in the CLIP text encoder: 1) Normal Text Prompt which describes the object in a typical, normal state (e.g., A photo of a normal screw). 2) Abnormal Text Prompt which describes the object in a defective state (e.g., A photo of a defective screw). 3) No State Text Prompt: A generic description of the object without any state-specific information (e.g., A photo of a screw). The process is:
\begin{equation}
    F_v = Linear(Encoder_v(I)),
    \label{equ: visual feature}
\end{equation}
\begin{equation}
    F_t^{\{n,a,s\}} = Linear(Encoder_t(T^{\{n,a,s\}})),
    \label{equ: textual features}
\end{equation}
$Linear(\cdot)$ and $Encoder(\cdot)$ denote the linear adapter and feature encoder, respectively. $F_t^{\{n,a,s\}}$ represents normal, abnormal and stateless textual embeddings, respectively. 

Once we obtain the stateless text embedding \(F_t^{s}\), our goal is to refine it with anomalous state that is implicitly presented in the image feature \(F_v\).
To rectify the stateless text embeddings with visual features, CMFR processes features in spatial-wise dimension, enabling better multi-modal feature interaction and state information insertion. The bi-modal inputs $\{F_{v,i}\in \mathbb{R}^{1\times D}\}_{i=1}^N$ and $F_t^{s}\in \mathbb{R}^{1\times D}$ are concatenated and embedded into two spatial weight maps: $\{M_{v,i}\in \mathbb{R}^{1\times D}\}_{i=1}^N$ and $\{M_{t,i}\in \mathbb{R}^{1\times D}\}_{i=1}^N$. The embedding operations are consisting of two convolution layers assembled with a RELU function and one linear projection adapter:
\begin{equation}
    F_i = Concate(F_{v,i}, F_t^{s}),
    \label{equ: concat}
\end{equation}
\begin{equation}
    \tilde{F_i} = Linear\bigr(Sigmoid(Conv(RELU(Conv(F_i))))\bigr),
    \label{equ: compute}
\end{equation}
\begin{equation}
    M_{v,i}, M_{t,i} = Spilt(\tilde{F_i}),
    \label{equ: split}
\end{equation}
where, $Conv(\cdot)$ and $Sigmoid(\cdot)$ denote convolution layer and sigmoid function, respectively.

The cross-modal feature rectification is then operated as:
\begin{equation}
    F_{t,i}^{s} = F_t^s + F_{v,i} \odot M_{v,i},
    \label{equ: rectify}
\end{equation}
where, $\odot$ denotes spatial-wise multiplication. Adding this residual to the vanilla prompt yields the rectified text feature. Then the newly injected features amplifies subtle features that distinguish a “normal” from an “abnormal” instance (scratches, dents, color shifts, etc.).

By integrating these steps, CMFR effectively enhances CLIP's ability to detect anomalies by refining the text embeddings to better capture anomalous state. This approach enhances the perception to anomalous regions and improves the robustness of anomaly detection on unseen data.

\subsection{Image-Level Text-on-Text Anomaly Scoring}
Traditional CLIP-based detectors compare class token $F_{v,1}$ with a static stateful text prompt, causing the score to be dominated by category semantics.  
In contrast, AFR-CLIP computes a text-on-text similarity, where the rectified embedding \(F_{t,i}^s\) is matched with two unified stateful prototypes:
\begin{equation}
    S_a = \cos(F_{t,1}^{s},\, F_t^{a}),
    \label{equ: sa}
\end{equation}
\begin{equation}
    S_n = \cos(F_{t,1}^{s},\, F_t^{n}),
    \label{equ: sn}
\end{equation}
and then converted to a probability via a softmax:
\begin{equation}
    P_{a}=\frac{\exp(S_a)}
    {\exp(S_a)+\exp(S_n)}.
    \label{equ: prob}
\end{equation}

\subsection{Pixel-Level Text-on-Text Anomaly Scoring}
For each patch feature $F_{v,i}\in\mathbb{R}^{i\times D}$ ($i=2,\dots,N$),  
we inject its anomalous state into the stateless prompt to obtain a patch-specific rectified text embedding:
\begin{equation}
F_{t,i}^s=F_t^{s} + F_{v,i} \odot M_{v,i}.
\label{eq:patch_calib}
\end{equation}
We then compare $F^s_{t,i}$ with the two unified prototypes:
\begin{equation}
S_{a,i}=\cos(F_{t,i}^s,F_t^{a}),
\end{equation}
\begin{equation}
S_{n,i}=\cos(F^s_{t,i},F_t^{n}),
\end{equation}
and convert them into an abnormality probability:
\begin{equation}
    P_{a,i}=\frac{\exp(S_{a,i})}
            {\exp(S_{a,i})+\exp(S_{n,i})}.
\label{eq:patch_prob}
\end{equation}
The set $\{P_{a,i}\}_{i=2}^{N}$ is then reshaped to $\mathbb{R}^{h\times w}$ and up-sampled to the input resolution $\mathbb{R}^{H\times W}$, yielding a dense anomaly heat map.
\begin{equation}
    HeatMap=Bilin\bigr(Reshape(\{P_{a,i}\}_{i=2}^{N})\bigr),
\label{eq:pixmap}
\end{equation}
where, $Bilin(\cdot)$ denotes bilinear interpolation.

\subsection{Self Prompting (SP)}
\label{subsection:SP}
CLIP’s ViT encoder excels at global reasoning, yet fine-grained industrial defects often manifest in local textures best captured by CNNs.  
SP insert corresponding CNN features into patch embeddings as visual prompt, fusing local and global context.
In detail, we employ $K$ linear projection adapters to transform the CNN features into visual prompts $P_v\in \mathbb{R}^{K\times D_v}$, aligning the feature dimensions across different model. The feed-forward process is:
\begin{equation}
    P_v = Concat\bigr([Linear_i({F}_{cnn})]_{i=1}^{K}\bigr),
    \label{equ：LN5}
\end{equation}
where $Concat(\cdot)$ denotes feature concatenation.

The visual prompt is designed for individual image, and a unified visual prompt is also essential for better generalization. To achieve this, we create $K$ learnable prompt tokens $ P_l\in \mathbb{R}^{K \times D_v}$ and combine $P_l$ with $P_v$ to form the final visual prompt tokens, as formulated in Eq. (\ref{equ：pt}).
\begin{equation}
    \tilde{P_v} = P_v + P_l.
    \label{equ：pt}
\end{equation}

CNN excels at capturing fine-grained local details, whereas ViT leverages self-attention to effectively model global context by allowing interactions across the entire token sequence. To harness these complementary strengths, we integrate the local and global features from the two frameworks through the self-attention mechanism inherent to ViT. As illustrated in Fig. \ref{fig:Agg-CLIP}, the CLIP image encoder contains 24 transformer layers, which we partition equally into four stages. To achieve effective fusion of local and global information, we insert the visual prompts exclusively to the first stage ($ST_1$).

More concretely, $ST_1$ comprises six transformer layers. To preserve the original contextual information from the input image, the first layer remains unmodified. For the 2nd to 6th layers, we replace the last $K$ tokens of the transformer layer’s output $F_v\in\mathbb{R}^{K\times D_v}$ with the modified visual prompt tokens $\tilde{P_v}$. Formally, this process is expressed as:
\begin{equation}
\label{eq:tf}
    L_{v_1} = TF_1(I),
\end{equation}
\begin{equation}
\label{eq:concat}
    L_{v_i} = TF_i\bigl(Concat(L_{v_{i-1}}^{1:N-K}, \tilde{P_v})\bigr),i=2,\ldots,6,
\end{equation}
where $TF_i(\cdot)$ denotes the $i$-th transformer layer and $L_{v_i}$ means the visual features of $i$-th layer in stage 1.
The visual prompts yield robust representations, substantially improving the model’s precision and robustness in detecting subtle anomalies. It's noting that the final result is obtained by averaging the outputs from all four stages.

\subsection{Multi-Patch Feature Aggregation (MPFA)}
\label{subsection:MPFA}
Although SP enhances local feature details, it faces challenges in capturing small defects due to its limited receptive field. To overcome this limitation, MPFA aggregates multi-scale information from neighboring patches, ensuring that small defects are effectively captured.

Given the adapted patch embeddings \(F_{v}^i\in\mathbb{R}^{N\times D_v}\) obtained at the \(i\)-th stage \(ST_i\), we first reshape these embeddings into a spatial grid format $\tilde{F}_{v}^i\in\mathbb{R}^{\sqrt{N}\times\sqrt{N}\times D_v}$.
\begin{equation}
    \tilde{F}_{v}^i=\mathrm{Reshape}(F_{v}^i).
\end{equation}
Next, we perform adaptive average pooling (AAP) within a \(m\times m\) local neighborhood around each patch (defaulting to \(m=3\)), thereby effectively integrating the surrounding spatial context into each patch’s representation:
\begin{equation}
\tilde{F}_{v_i}^m=\mathrm{AAP}(\tilde{F}_{v_i})\in\mathbb{R}^{\sqrt{N}\times\sqrt{N}\times D_v}.
\end{equation}
Finally, the enhanced feature maps are reshaped back into their original patch-level format for subsequent processing:
\begin{equation}
F_{v_i}^m=\mathrm{Reshape}(\tilde{F}_{v_i}^m)\in\mathbb{R}^{N\times D_v}.
\end{equation}

\begin{table*}[t]
  \centering
  \caption{Image-level performance comparisons of the ZSAD methods on the industrial and medical datasets. The best performance is in \textbf{bold} and the second-best is \underline{underlined}.}
  \setlength{\tabcolsep}{1mm}  
  \small  
  \begin{tabular}{lccccccccccc}
    \toprule[1.5pt]
    \multirow{2}{*}{Domain} & \multirow{2}{*}{Dataset} & \multicolumn{2}{c}{WinCLIP} & \multicolumn{2}{c}{APRILGAN} & \multicolumn{2}{c}{AnomalyCLIP} & \multicolumn{2}{c}{AdaCLIP} & \multicolumn{2}{c}{\textbf{AFR-CLIP (Ours)}} \\
    \cmidrule(lr){3-4}\cmidrule(lr){5-6}\cmidrule(lr){7-8}\cmidrule(lr){9-10}\cmidrule(l){11-12}
    & & AUROC & max-F1  & AUROC & max-F1  & AUROC & max-F1  & AUROC & max-F1  & AUROC & max-F1 \\
    \midrule
    \multirow{5}{*}{Industrial} 
    & VisA    & ( 78.1 ~, & 79.0 )  &( 81.7 ~,  &80.7 ) &( 82.1 ~,  & 80.4 ) &( \underline{85.8} ~,  &\underline{83.1} )  &( \textbf{87.0} ~, & \textbf{84.7} )  \\
    
    & BTAD    & ( 68.2 ~, &67.6 )  &( 85.2 ~,  &82.0 ) &( 88.3 ~,  & 83.8 ) &( \underline{88.6} ~,  &\underline{88.2} )  &( \textbf{95.8} ~, & \textbf{93.2} )  \\

    & MPDD    & ( 61.4 ~, &77.5 )  &( 63.0 ~,  &77.0 ) &( \underline{77.0} ~,  & 80.4 ) &( 76.0 ~,  &\underline{82.5} )  &( \textbf{77.9} ~, & \textbf{84.5} )  \\
    
    & MVTec AD & ( 91.8 ~,  & \textbf{92.9} )  & ( 82.3 ~, & 88.9 )  & ( 91.5 ~, & \underline{92.7} ) & ( 89.2 ~,  & 90.6 )  & ( \textbf{93.4} ~,  & 92.3 ) \\
    \cmidrule(l){2-12}
    
    &\textit{Average} & ( 74.9 ~, &79.3 )  &( 78.1 ~,  &82.2 ) &( 84.8 ~,  & 84.3 ) &( \underline{84.9} ~,  &\underline{86.1} )  &( \textbf{88.5} ~, & \textbf{88.7} )  \\
    
    \midrule
    \multirow{4}{*}{Medical}    
    & Head CT    & ( 81.8 ~, & 79.8 )  &( 87.5 ~,  & 88.0 ) &( \underline{93.4} ~,  & \textbf{90.8} ) &( 91.8 ~,  & 84.1 )  &( \textbf{94.5} ~, & \underline{89.6} )  \\
    
    & Brain MRI & ( 86.6 ~, & 86.3 )  &( 88.0 ~,  & 83.1 ) &( 90.3 ~,  & \underline{90.2} ) &( \underline{93.5} ~,  & 89.7 )  &( \textbf{93.6} ~, & \textbf{91.8} )  \\
                                
    & Br35H    & ( 80.5 ~, & 74.4 )  &( 91.5 ~,  & 84.9 ) &( \underline{94.6} ~,  & \underline{89.1} ) &( 92.3 ~,  & 85.3 )  &( \textbf{95.1} ~, & \textbf{89.9} )  \\

    \cmidrule(l){2-12}
    
    &\textit{Average} & ( 83.0 ~, & 80.2 )  &( 89.0 ~,  & 85.3 ) &( \underline{92.8} ~,  & \underline{90.0} ) &( 92.5 ~,  & 86.4 )  &( \textbf{94.4} ~, & \textbf{90.4} )  \\
    \bottomrule[1.5pt]
  \end{tabular}
  \label{tab:image-level}
\end{table*}

\begin{figure}[t]
	\centering
	\includegraphics[width=0.48
    \textwidth]{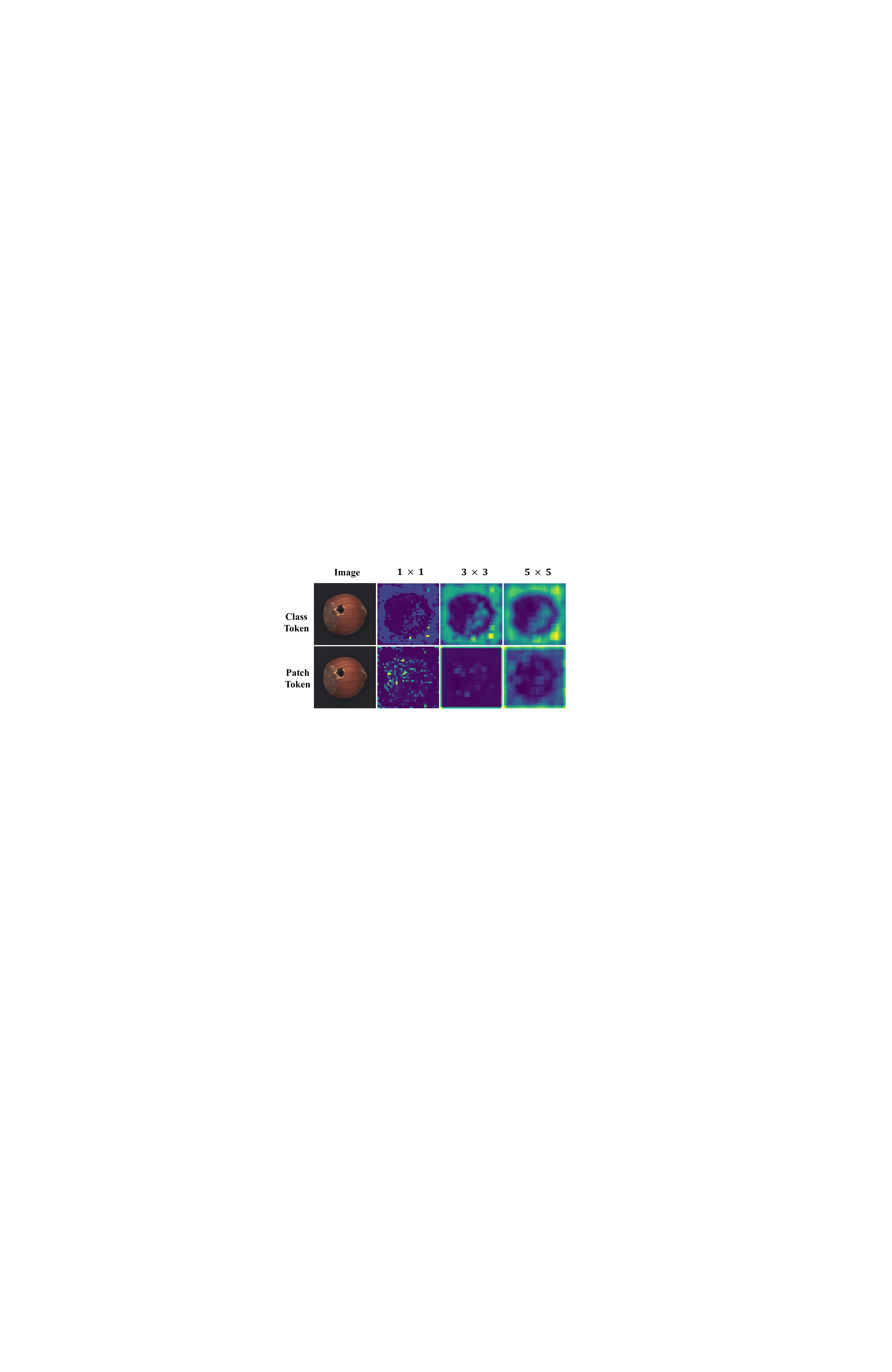} %
	\caption{Visualization of aggregated feature maps}
	\label{fig:agg}
\end{figure}

As illustrated in Fig.~\ref{fig:agg}, MPFA visibly enhances the clarity and coherence of feature maps at patch boundaries. This refined spatial continuity allows the model to capture more detailed contours and detect subtle defects more accurately.


\begin{table*}[t]
  \centering
  \caption{Pixel-level performance comparisons of the ZSAD methods on the industrial and medical datasets. The best performance is in \textbf{bold} and the second-best is \underline{underlined}.}
  \setlength{\tabcolsep}{1mm}  
  \small  
  \begin{tabular}{lccccccccccc}
    \toprule[1.5pt]
    \multirow{2}{*}{Domain} & \multirow{2}{*}{Dataset} & \multicolumn{2}{c}{WinCLIP} & \multicolumn{2}{c}{APRILGAN} & \multicolumn{2}{c}{AnomalyCLIP} & \multicolumn{2}{c}{AdaCLIP} & \multicolumn{2}{c}{\textbf{AFR-CLIP (Ours)}} \\
    \cmidrule(lr){3-4}\cmidrule(lr){5-6}\cmidrule(lr){7-8}\cmidrule(lr){9-10}\cmidrule(l){11-12}
    & & AUROC & max-F1  & AUROC & max-F1  & AUROC & max-F1  & AUROC & max-F1  & AUROC & max-F1 \\
    \midrule
    \multirow{5}{*}{Industrial} 
    & VisA    & ( 79.6 ~, &14.8 )  &( 95.2 ~,  &32.3  ) &( 95.5 ~,  & 28.3 ) &( \underline{95.5} ~,  & \underline{37.7} )  &( \textbf{96.0} ~, & \textbf{38.1} )  \\
    
    & BTAD    & ( 72.6 ~, &18.5 )  &( 89.5 ~,  &38.4 ) &( \underline{94.2} ~,  & 49.7 ) &( 92.1 ~,  &\underline{51.7} )  &( \textbf{95.0} ~, & \textbf{52.1} )  \\

    & MPDD    & ( 71.2 ~, &15.4 )  &( 95.1 ~,  &30.6 ) &( \underline{96.5} ~,  & 34.2 ) &( 96.1 ~,  &\underline{34.9} )  &( \textbf{97.1} ~, & \textbf{35.6} )  \\
    
    & MVTec AD & ( 85.1 ~, & 31.6 )  &( 83.7 ~,  & 39.8 ) &( \underline{91.1} ~,  & 39.1 ) &( 88.7 ~,  & \underline{43.4} )  &( \textbf{92.3} ~, & \textbf{47.9} ) \\
    \cmidrule(l){2-12}
    
    &\textit{Average} & ( 77.1 ~, &20.1 )  &( 90.9 ~,  &35.3 ) &( \underline{94.3} ~,  & 37.5 ) &( 93.1 ~,  &\underline{41.9} )  &( \textbf{95.1} ~, & \textbf{43.4} )  \\
    
    \midrule
    \multirow{5}{*}{Medical}    
    & ISIC    & ( 83.3 ~, & 48.5 )  &( 85.8 ~,  & 67.3 ) &( 89.7 ~,  & 70.6 ) &( \underline{90.3}~,  & \underline{72.6} )  &( \textbf{90.6} ~, & \textbf{75.1} )  \\
    
    & ColonDB & ( 70.3 ~, & 19.6 )  &( 78.4 ~,  & 33.2 ) &( 81.9 ~,  & \textbf{37.3} ) &( \underline{82.6} ~,  & 36.1 )  &( \textbf{83.3} ~, & \underline{36.4} )  \\
                                
    & ClinicDB & ( 51.2 ~, & 24.4 )  &( \underline{83.2} ~,  & \underline{42.3} ) &( 82.9 ~,  & 42.1 ) &( 82.8 ~,  & 40.9 )  &( \textbf{87.5} ~, & \textbf{45.8} )  \\

    & TN3K & ( 70.7 ~, & 30.0 )  &( 74.4 ~,  & 39.7 ) &( \textbf{81.5} ~,  & \textbf{47.9} ) &( 76.8 ~,  & 40.7 )  &( \underline{78.2} ~, & \underline{46.7} )  \\

    \cmidrule(l){2-12}
    
    &\textit{Average} & ( 68.9 ~, & 30.6 )  &( 80.5 ~,  & 45.6 ) &( \underline{84.0} ~,  & \underline{49.5} ) &( 83.1 ~,  & 47.6 )  &( \textbf{84.9} ~, & \textbf{51.0} )  \\
    \bottomrule[1.5pt]
  \end{tabular}
  \label{tab:pixel-level}
\end{table*}

\section{Experiments}
\subsection{Experimental Setup}
\textbf{Datasets.} Following the existing methods, we conduct experiments across both industrial and medical domains to validate the performance of our model AFR-CLIP. For the industrial domain we use MVTec AD \cite{mvtec}, VisA \cite{visa}, BTAD~\cite{btad} and MPDD~\cite{mpdd}, while for the medical domain we adopt the Brain MRI~\cite{brainmri}, Head CT~\cite{headct}, Br35H~\cite{br35h}, ISIC~\cite{isic}, ColonDB~\cite{colon}, ClinicDB~\cite{clinic} and TN3K~\cite{tn3k}. 
Because the object categories in VisA do not overlap with those in the other datasets, we follow a cross-dataset protocol: the model is trained on VisA and tested on other datasets, whereas for VisA evaluation we employ the model trained on MVTec AD. 

\textbf{Evaluation Metrics.} Following the prior works in zero-shot anomaly detection, we employ Area Under the Receiver Operating Characteristic Curve (AUCROC) and the maximum F1 score (max-F1) to compare the performance with other ZASD approaches in anomaly detection and localization. To provide a more comprehensive assessment, we also present the mean performance over every domain.

\textbf{Implementation Details.} Our experiments select the pre-trained CLIP with ViT-L-14@336 by OpenAI as default backbone. 
All images are resized to $518 \times 518$ during both training and inference. 
The 24 transformer layers in the CLIP image encoder are grouped into four stages, and patch embeddings are taken from the 6th, 12th, 18th, and 24th layers. 
For self prompting, we set the number of adapters to \(K = 5\), resulting in a visual-prompt size of \(5 \times D\). 
For multi-patch feature aggregation, the neighborhood window is fixed at \(m = 3\). 
The network is trained for 100 epochs with a batch size of 4 using the Adam optimizer. 
The initial learning rate is \(0.001\) and decays according to \(\mathrm{lr} = 0.001 / (\text{epoch} + 1)\).
All experiments are conducted on a single NVIDIA RTX A6000 GPU (48\,GB).

\textbf{Comparison Methods.} To better evaluate our model, we compare our proposed AFR-CLIP with two kinds of methods: training-free and training with auxiliary data methods. For the training-free methods, we select WinCLIP \cite{winclip} for comparison, the first representative CLIP-based method and requesting no auxiliary data for training. For the second kind of methods, we select APRIL-GAN \cite{aprilgan}, AdaCLIP \cite{adaclip} and AnomalyCLIP \cite{anomalyclip} for comparison, training on auxiliary data and testing on novel objects.

\subsection{Comparisons with State-of-the-Art Methods}
\textbf{Quantitative Results.} 
As reported in Tab.~\ref{tab:image-level},
AFR-CLIP achieves superior image-level performance on both industrial and medical datasets. In the industrial domain, AFR-CLIP leads with 87.0\% AUROC and 84.7\% max-F1 on the VisA dataset, surpassing the best competing method by 5.3\% AUROC and 4.0\% max-F1. Similarly, on the MVTec AD dataset, AFR-CLIP improves performance by 1.6\% AUROC, reaching 93.4\% compared to the previous best method. In the medical domain, AFR-CLIP shows remarkable improvements, especially on the Head CT dataset with 94.5\% AUROC and 89.6\% max-F1, outperforming AdaCLIP by 2.7\% in AUROC and 5.5\% in max-F1. 

At the pixel-level, AFR-CLIP achieves the best anomaly localization results across all datasets. On the industrial datasets, AFR-CLIP outperforms the second best by up to 1.0\% in P-AUC, reaching 96.0\% on VisA and 92.3\% on MVTec AD. The model's ability to localize complex anomalies such as small cracks and texture defects is demonstrated by a 3.9\% improvement in pixel-level accuracy on MVTec AD compared to AdaCLIP.
In the medical domain, AFR-CLIP also shows consistent improvements over many datasets, with 90.6\% P-AUC on ISIC, surpassing the next best method by 2.3\%.
On ClinicDB and TN3K datasets, AFR-CLIP likewise achieves the highest P-AUC scores.
These margins validate the core motivation: by rectifying the stateless prompt with image-specific cues and replacing image-text with text-on-text similarity, AFR-CLIP suppresses category semantics and elevates anomaly-state signals, leading to consistent gains across domains.

\begin{figure}[htbp]
	\centering
	\includegraphics[width=0.48\textwidth]{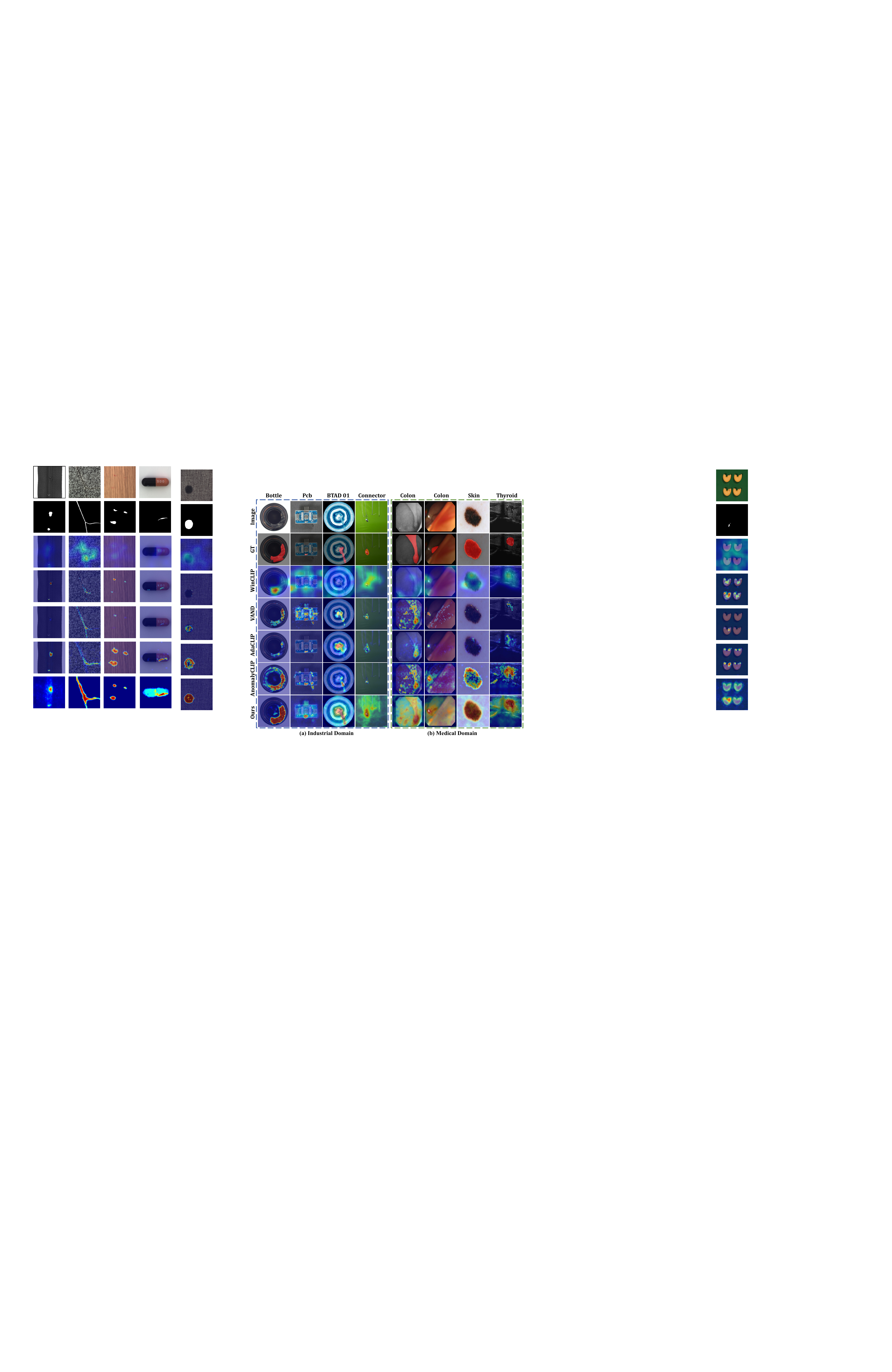} %
	\caption{Visualization of ZSAD methods. (a) and (b) present the localization results across both industrial and medical domains.}
	\label{fig:visual}
\end{figure}

\textbf{Qualitative Results} across various datasets are presented in Fig. \ref{fig:visual}.
In the industrial domain, we observe small defects like the scratch in the pcb and complex defects like large cracks in the bottle. The ground truth for these anomalies are fine-grained, and our AFR-CLIP consistently outperforms other methods in localizing their contours. In categories like connector and bottle, AFR-CLIP’s pixel-level results clearly demonstrate its superiority to detect and localize defects accurately.
On the medical side, as seen in Fig.~\ref{fig:visual}(b), AFR-CLIP also provides more precise localization of anomalies compared to other methods, producing masks that are much closer to the ground truth. In particular, for the skin, colon, and thyroid categories, our method performs significantly better on this task, yielding more accurate and complete segmentations with fewer false negatives.
The visual improvements further confirm that our cross-modal anomaly feature rectification direct the model’s attention to anomalous state, thereby enabling more complete localization of defective regions than prior CLIP-based methods.

\subsection{Ablation Study}

As reported in Tab. \ref{tab:ablation}, we conduct ablation studies on MVTec AD to assess the influence of every component in AFR-CLIP.
In Tab.~\ref{tab:CFFA}, we conduct an ablation study on the the impact of the different stages where SP module is employed. Moreover, we perform the degree analysis by varying the size of neighboring patches aggregated during the MPFA process as reported in Tab. \ref{tab:MPFA}.

\begin{table}[t]
\centering
\small
\caption{Ablation Results of AFR-CLIP's Components.}
\label{tab:ablation}
\resizebox{1\linewidth}{!}{
\begin{tabular}{ccccccc}
\toprule
Num.& CMFR & SP & MPFA & Image‐level & Pixel‐level\\
\midrule
  1.& &  &  & 89.1 & 88.4\\
  2.& & \ding{52} & \ding{52} & 90.7 (+1.6) & 90.1 (+1.7)\\
  3.& \ding{52} &  & & 92.1 (+3.0) & 91.2 (+1.8)\\
  4.& \ding{52} & \ding{52} &  & 92.5 (+3.4) & 91.6 (+3.2)\\
  5.& \ding{52} &  & \ding{52} & \underline{92.9} (+3.8) & \underline{91.8} (+3.4)\\
  6.& \ding{52} & \ding{52} & \ding{52} & \textbf{93.4 (+4.3)} & \textbf{92.3 (+3.9)}\\
\bottomrule
\end{tabular}}
\end{table}

\textbf{Ablation for Cross-Modal Feature Rectification (CMFR) module.} As reported in Tab. \ref{tab:ablation}, the performance gain from the first row to the third row demonstrates superiority of CMFR module. We measure text-on-text similarity while existing methods obtain the anomaly map by measuring the cosine similarity between visual and textual features. With CMFR, our text-on-text anomaly scoring gains a performance improvement, surpassing existing methods by 3.0\% and 1.8\% of image-level and pixel-level AUCROC, respectively. After CMFR, the anomalous information implied in the image is injected into stateless text prompt, yielding a stateful textual embedding. In this way, the model's attention is directed more to the anomalous state rather than the object category. 

\begin{table}[t]
\centering
 
\caption{Ablation on the Employed Stages of SP.}
\label{tab:CFFA}
\begin{tabular}{ccc}
\toprule
Stage & Image-Level & Pixel-Level                   \\
\midrule
$1$ & \underline{93.4} & \textbf{92.3}  \\ 
$1-2$ & \textbf{93.5} & \underline{92.2} \\ 
$1-3$ & 92.8 & 91.9  \\ 
$1-4$ & 93.0 & 92.1  \\ 
\bottomrule
\end{tabular}
\end{table}

\begin{table}[t]
\centering
\caption{Ablation Results of $P_v$ and $P_l$ in MPFA.}
\label{tab:vl}
 
\begin{tabular}{cccc}
\toprule
$P_v$ & $P_l$ & Image-Level & Pixel-Level                   \\
\midrule
& & 92.9 & 91.8  \\ 
\ding{52}& & 93.1 & \underline{92.1}  \\ 
&\ding{52} & \underline{93.2} &  92.0 \\ 
\ding{52}&\ding{52} & \textbf{93.4} & \textbf{92.3}  \\ 
\bottomrule
\end{tabular}
\end{table}

\begin{table}[t]
\centering
\caption{Ablation Results of Different Values of $m$ in MPFA.}
\label{tab:MPFA}
 
\begin{tabular}{ccc}
\toprule
Scale & Image-Level & Pixel-Level                   \\
\midrule
$1\times 1$ & 92.5 & 91.6  \\ 
$3\times 3$ & \textbf{93.4} & \textbf{92.3}  \\ 
$5\times 5$ & \underline{93.2} & \underline{92.1}  \\ 
\bottomrule
\end{tabular}
\end{table}

\textbf{Ablation for self prompting (SP) module.} To assess the impact of the SP module, we conduct an ablation experiment comparing the performance of AFR-CLIP with and without SP on the MVTec AD dataset. Comparing the third and fourth rows in Tab. \ref{tab:ablation}, the performance on both detection and localization increases 0.4\% AUCROC. With SP module, the model can better capture detailed local context. In this way, the model effectively aligns features from both CNN and ViT, enabling the fusion of local and global representations. 
Additionally, we evaluate the impact of applying Self-Prompting (SP) at different stages in Tab. \ref{tab:CFFA}. Notably, when applied to the 2nd to 4th stages, there is no significant performance improvement. This may be attributed to the fact that feature interactions have already been done in the first stage, and applying it multiple times may lead to overfitting. To optimize computational efficiency and obtain the best performance, we apply SP only on the first stage. Moreover, we explore the importance of both the image-specific prompt $P_v$ and the unified prompt $P_l$ in Tab.~\ref{tab:vl}. Both of these prompts are crucial, with $P_v$ enhancing image-specific features and $P_l$ encapsulating shared information.

\textbf{Ablation for multi-patch feature aggregation (MPFA) module.}
To assess the importance of Adaptive average pooling over $m\times m$ neighborhood patch embeddings, where $m$ is set to 3 in default. we conduct corresponding ablation study as shown in the fifth row of Tab. \ref{tab:ablation}. When it comes to small and complex defects like crack in pill and bent wire in cable, the MPFA module can strengthen the network’s perception of large-scale features, which results in an increase of 0.8\% in I-AUC and 1.6\% in P-AUC.
Furthermore, we search for the best value of pooling size $m$ by comparing the results of \{1, 3, 5\} reported in Tab. \ref{tab:MPFA}. The visualization of patch embeddings aggregated by different sizes is illustrated in Fig. \ref{fig:agg}. Properly aggregating the neighborhood patch embeddings can contributes to the bigger scale and complex defects, while the feature will be vague if $m$ is too big.

\section{Conclusion}
Zero-shot anomaly detection remains challenging because conventional CLIP-based methods prioritize object category and therefore overlook the subtle state cues that distinguish normal from defective instances.
In this paper, we propose AFR-CLIP, a novel zero-shot anomaly detection framework that departs from traditional image-text matching by 1) performing stateless-to-stateful cross-modal rectification to inject defect cues into the textual embeddings and 2) computing anomaly map via text-on-text similarity, thus shifting the focus from object category (e.g., screw, bottle) to anomalous state (normal or abnormal).
Further, the self prompting (SP) module enhances feature representation by effectively combining global and local context, while the multi-patch feature aggregation (MPFA) module further refines anomaly detection by improving the model's perception to complex and small defects. Extensive experiments on both industrial and medical datasets validate the superiority of AFR-CLIP in zero-shot anomaly detection, showcasing its ability to generalize robustly across diverse domains.

\bibliography{aaai2026}
\clearpage

\end{document}